\documentclass[letterpaper, 10 pt, conference]{ieeeconf}  

\IEEEoverridecommandlockouts                              

\usepackage{cite}
\usepackage{amsmath,amssymb,amsfonts}
\usepackage{algorithmic}
\usepackage{graphicx}
\usepackage{textcomp}
\usepackage{xcolor}
\usepackage{authblk}
\usepackage{wrapfig}
\usepackage{float}

\usepackage{balance}
\usepackage{lastpage}

\title{\LARGE \bf
Detection of E-scooter Riders in Naturalistic Scenes
}

\author{Kumar Apurv$^{1}$, Renran Tian$^{{2}^\dag}$, and Rini Sherony$^{3}$
\thanks{*This work was supported by the Collaborative Safety Research Center (CSRC), Toyota Motor North America}
\thanks{$^\dag$Corresponding author.  E-mail:  rtian@iupui.edu.}%
\thanks{$^{1}$Kumar Apurv is with the Department of Computer \& Information Science, Indiana University-Purdue University Indianapolis (IUPUI), SL 280, 723 W. Michigan Street, Indianapolis, IN 46202, USA.}%
\thanks{$^{2}$Renran Tian is with the Department of Computer Information \& Graphics Technology, Purdue School of Engineering and Technology, Indiana University-Purdue University Indianapolis (IUPUI), ET 301, 799 West Michigan Street, Indianapolis, IN 46202, USA.}%
\thanks{$^{3}$Rini Sherony is with the Collaborative Safety Research Center (CSRC), Toyota Motor North America, 1555 Woodridge Ave, Ann Arbor, MI 48105, USA.}%

}

\begin{document}

\maketitle
\thispagestyle{empty}
\pagestyle{empty}

\begin{abstract}

E-scooters have become ubiquitous vehicles in major cities around the world. The numbers of e-scooters keep escalating, increasing their interactions with other cars on the road. An e-scooter rider’s normal behavior varies enormously to other vulnerable road users. This situation creates new challenges for vehicle active safety systems and automated driving functionalities, which require the detection of e-scooter riders as the first step. To our best knowledge, there is no existing computer vision model to detect these e-scooter riders. This paper presents a novel vision-based system to differentiate between e-scooter riders and regular pedestrians and a benchmark data set for e-scooter riders in natural scenes. We propose an efficient pipeline built over two existing state-of-the-art convolutional neural networks (CNN), You Only Look Once (YOLOv3) and MobileNetV2. We fine-tune MobileNetV2 over our dataset and train the model to classify e-scooter riders and pedestrians. We obtain a recall of around 0.75 on our raw test sample to classify e-scooter riders with the whole pipeline. Moreover, the classification accuracy of trained MobileNetV2 on top of YOLOv3 is over 91\%, with precision and recall over 0.9.

\end{abstract}

\section{INTRODUCTION}

E-scooters have taken major cities across the world by storm.  They are convenient, cost-efficient, and environment friendly, and above all, they are fun to ride. Studies show that there has been an exponential increase in the rise of e-scooter riders. But with this invasion, there also comes a spike in injuries and accidents. The Journal of the American Medical Association (JAMA) shows that after the e-scooter sharing companies took off in 2018, the annual hospital admissions related to e-scooters climbed by 365\% in North America~\cite{namiri2020electric}. Though the accidents are concerning, the e-scooters are still considered a noble attempt to make the big cities comfortable to navigate as an agile micro-mobility tool. Much effort is needed to improve overall safety related to e-scooters and their interactions with cars and pedestrians in the naturalistic road environment.

Computer vision models to detect e-scooter riders can help understand their behavior and help predict and mitigate potential accidents. In the era of deep learning, we have seen some astonishing progress in the detection of pedestrians. Today, state-of-the-art algorithms can detect these e-scooter riders but classify them as pedestrians. These algorithms cannot determine whether the detected person is riding an e-scooter or not. The expected behavior of an e-scooter rider on the road is very different from regular pedestrians, thus making it crucial to differentiate such riders from other pedestrians on the road. 

In this paper, we present a classification approach to identify these e-scooter riders. Since the e-scooters had come to light only a few years ago, there are no benchmark data-sets today to put the state-of-the-art detection models into effect. Deep learning requires large data-sets, and therefore we create an original benchmark data-set to train the classification model. The proposed vision-based system to detect the e-scooter riders can serve the autonomous driving community and behavioral analysis domains.

We have organized the paper in the following manner. In section II, we go over some remarkable achievements in pedestrian detection and generic object detection. We also discuss similar works such as vision models to distinguish cyclists. In section III, we go step-by-step through the proposed algorithm to detect the e-scooter riders, our model’s pipeline, and architecture. In section IV, we discuss the data collection and data-set creation process. Here, we also establish our training process and our optimization techniques. Finally, in section V, we discuss our model’s performance and share some visual results.


\section{RELATED WORK}

Over the last decade, we have seen astonishing improvements in the detection of pedestrians with the help of convolutional neural networks. Sermanet et al. introduced the ConvNet model in 2013 to produce competitive results on major pedestrian detection benchmarks~\cite{sermanet2013pedestrian}. As object detection algorithms like R-CNN~\cite{girshick2014rich}, Fast R-CNN~\cite{Girshick_2015_ICCV}, and Faster R-CNN~\cite{ren2015faster} got introduced, we started seeing more improvement in the detection of pedestrians in natural scenes. Some interesting works related to this are the algorithm proposed in ~\cite{li2017scale} uses a divide-and-conquer method over built-in subnetworks. An algorithm proposed in ~\cite{zhang2017pedestrian} presents a region proposal network and K-means cluster analysis for extraction. In 2016, Redmon and others introduced YOLO~\cite{redmon2016you}, a new approach to object detection by uniting feature extraction and localization into a single block. The authors released multiple versions of YOLO in recent years, with YOLOv3~\cite{redmon2018yolov3} providing a significant improvement over the initial version concerning performance.   

\begin{figure*}[h]
    \begin{center}
    \includegraphics[scale=0.42]{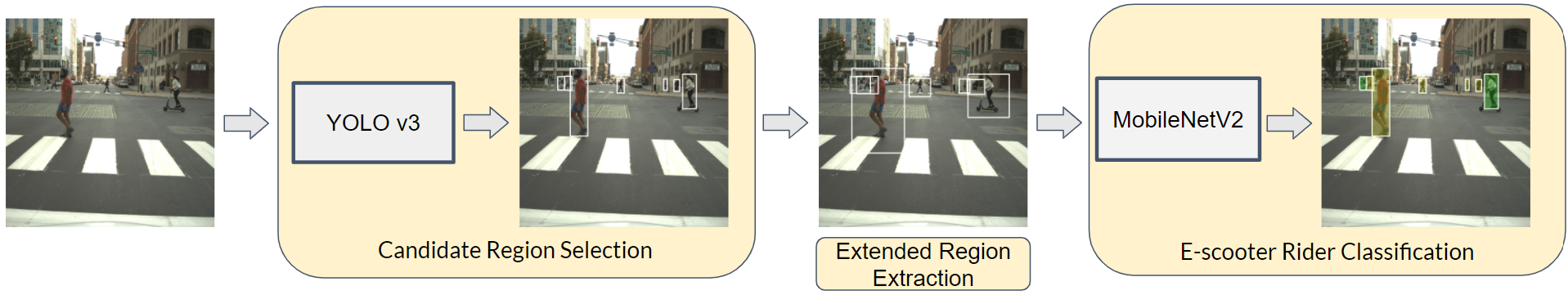}
    \caption{Model Pipeline: We divide the working of this model into three major steps. Firstly, we use a pre-trained model of YOLOv3 to extract the bounding boxes for persons in the image. In the second step, we update the bounding box values to account for larger boxes to encapsulate the e-scooter pixels if the person is riding. We use these extended bounding boxes to extract different image segments and pass them to the prediction model. In the last step, we use our trained version of MobileNetV2 to perform binary classification on these extracted regions to predict e-scooter riders.}
    \label{ModelArchitecture}
    \end{center}
\end{figure*}

Other state-of-the-art networks like MobileNetV2~\cite{sandler2018mobilenetv2} maintain representational power of the network using lightweight depthwise convolutions and exhibit excellent performance on image classification. Many authors have trained MobileNetV2 to achieve highly accurate models to perform several image classification tasks. In work by Qian et al.~\cite{xiang2019fruit}, they use MobileNetV2 as the base models to develop an algorithm through transfer learning to perform image classification on fruit images, with accuracy over 85\%.

Although we have come a long way with pedestrians’ detection and generic object detection, only a few detection models exist for cyclists and none for e-scooter riders. A framework proposed by Li et al.~\cite{li2016new, li2016unified} in 2016-17 introduced a new benchmark dataset and a detection model based on Fast R-CNN. Another deep neural network model established in 2019~\cite{wang2019pedestrian} proposes a system to discriminate between cyclist and pedestrian targets. Moreover, Saleh et al. proposed a Faster R-CNN for detection and localization for cyclists in depth images~\cite{saleh2017cyclist}.


\section{E-SCOOTER RIDER RECOGNITION ALGORITHM}

We create a deep learning model that takes video frames as input and determines whether the people in the frames are riding or not riding an e-scooter. Our model outputs bounding boxes around people with a prediction value to denote if they are riding an e-scooter or not. As shown in Fig.~\ref{ModelArchitecture}, we break down our algorithm into three different parts as discussed below:

\subsection{Candidate Region Selection}

YOLO or You Only Look Once ~\cite{redmon2018yolov3} is an end-to-end deep learning model developed by Redmon et al. for real-time object detection. Using a single neural network and regression technique, they predict the bounding boxes and the classification of objects. For our model, we use pre-trained model weights trained using DarkNet-53 on the MS COCO dataset~\cite{lin2014microsoft}. This dataset defines 80 different classes of objects, including persons, which is of our prime interest.

\begin{figure}[h]
    \centering
    \includegraphics[scale=0.16,trim={0 8.5cm 0 0},clip]{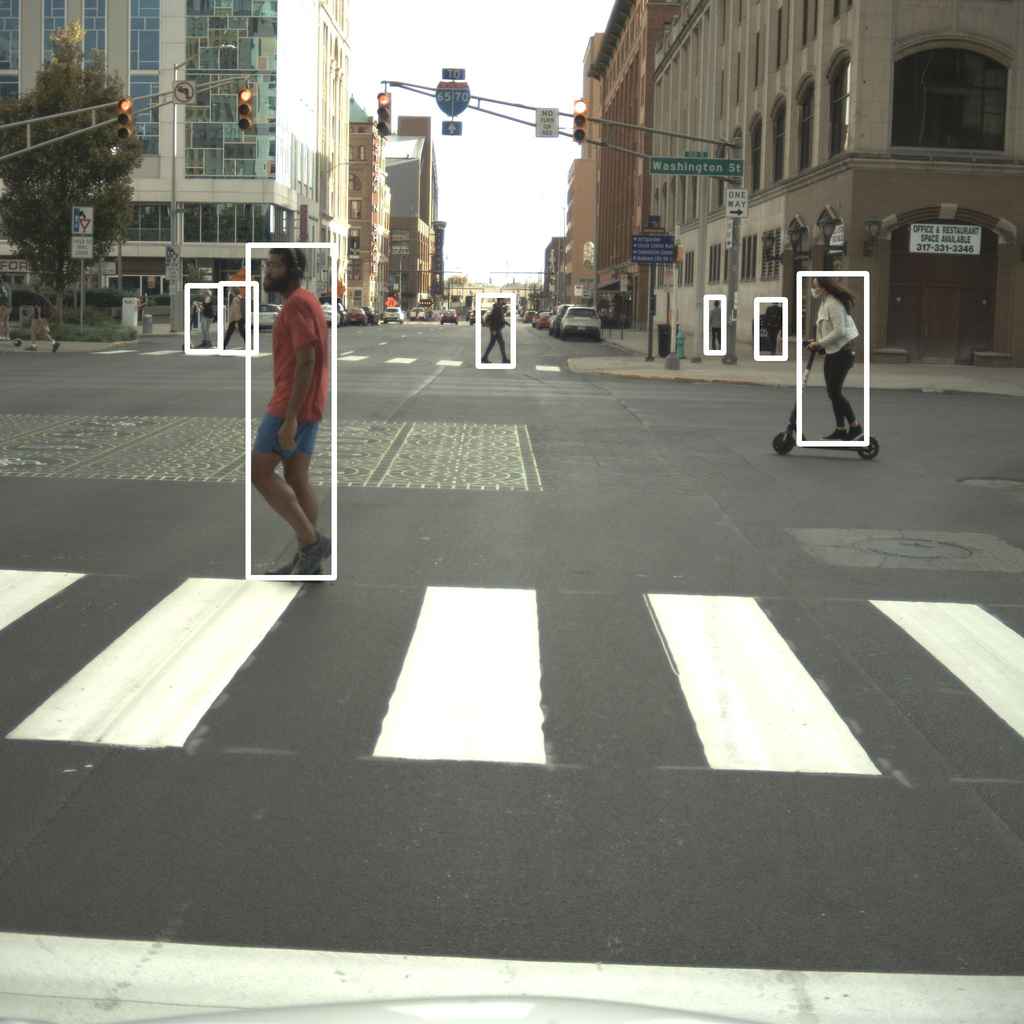}
    \caption{\textsf{Person detection with YOLOv3 for candidate region selection}}
    \label{fig:step1}
\end{figure}

YOLOv3 is a remarkable improvement over YOLO’s initial version using refined model architecture and the training process. YOLOv3 uses a very deep Darknet-53 as a feature extractor. Unlike the first version of YOLO, which used a Darknet version with only 19 layers, YOLOv3 uses Darknet-53, which uses 3x3 and 1x1 filters with residual networks after every few convolutional layers. Batch normalization layers and leaky ReLU activation follow each of these 53 convolutional layers. They avoid pooling as it often contributes to the loss of low-level features.

One frame at a time, the input feed is passed through YOLOv3 first. We resize the frames to the size of 416x416x3 before passing them into the YOLOv3 model. The YOLOv3 model detects people and outputs bounding box details for every person in the frame. We represent these bounding boxes with a four-dimensional vector, [x, y, w, h], where (x, y) is the top left corner of a bounding box, and w and h are the width and height of the box. So, in other words, we use YOLOv3 to output an nx4 dimensional vector, where n is the number of people detected in the frame.

\subsection{Extended Region Extraction} \label{process_step_2}

Fig.~\ref{fig:step1} shows that YOLOv3 outputs the bounding boxes for every person in the frame. We represent each bounding box as (x, y, w, h). Then, we process these boxes to obtain an enlarged bounding box to include the e-scooter pixels. Let us say that the enlarged bounding box’s dimensions are in the format (x’, y’, w’, h’). We enlarge the width on either side of the rectangle, and we broaden the height to 1.25 times the original. We generally find e-scooters in the lower section of the extracted image section. Therefore, we enlarge the box on all sides except the top. These steps ensure that the new bounding box encloses the e-scooter pixels if the person is riding one. As a result, the new dimensions of the box is as follows:
\begin{equation}
(x’, y’, w’, h’) = (x-w, y, 3w, h+h/4)
\end{equation}

\begin{figure}[h]
    \centering
    \includegraphics[scale=0.16,trim={0 8.5cm 0 0},clip]{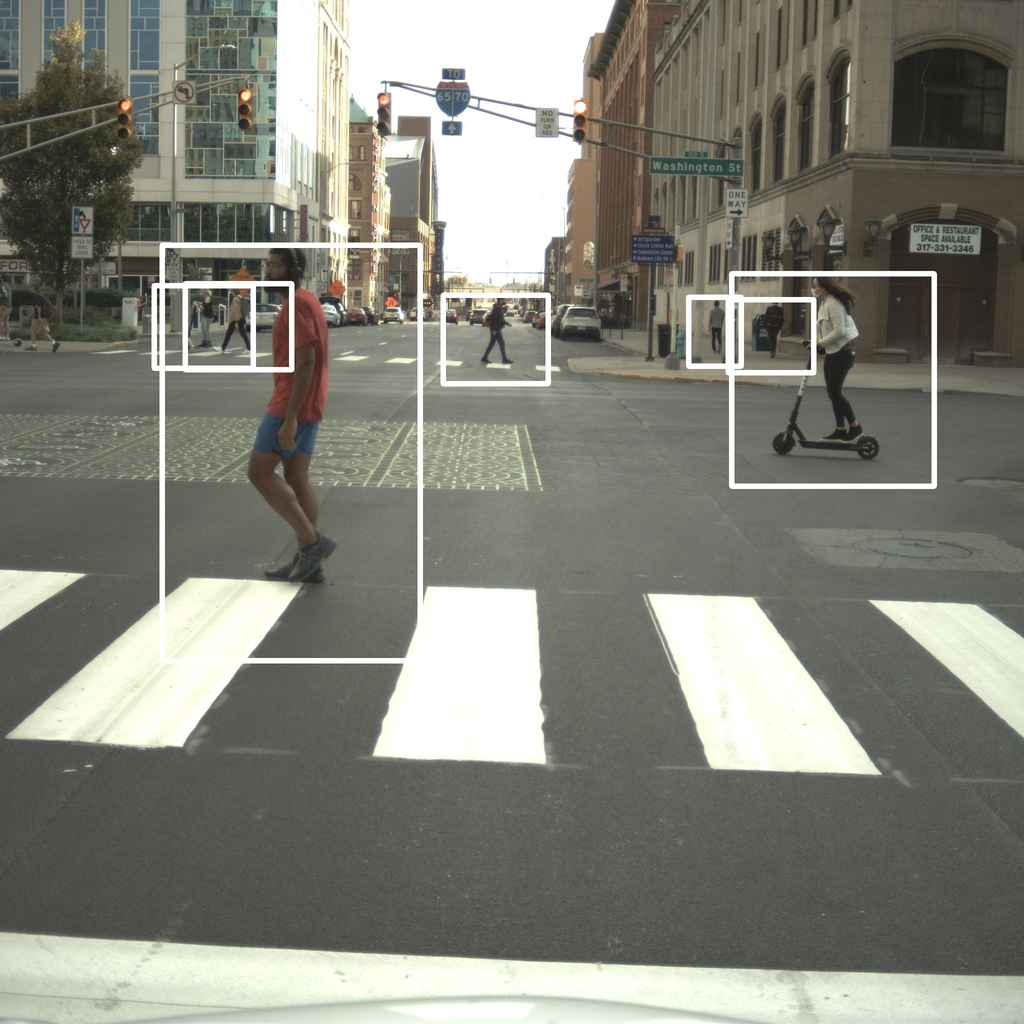}
    \caption{\textsf{Modified bounding boxes for extended region extraction}}
    \label{fig:step2}
\end{figure}

Based on this logic, we post-process this nx4 dimensional vector to account for larger bounding boxes around their corresponding detections, as shown in Fig.~\ref{fig:step2}. We then use the updated vector to extract pixel values for each detection from the original frame. Then, we represent each extracted segment using a 160x160x3 image vector, thus creating an (n, 160, 160, 3) tensor to represent all detections in the frame.

\subsection{E-scooter Rider Classification}

We then perform a single forward pass through our pre-trained version of MobileNetV2 to predict which of these n persons are riding an e-scooter. We train MobileNetV2 offline using our dataset to output a prediction value for each person in the range [0, 1]. We finally generate an (n,1) prediction tensor for each image. The deep CNNs take a massive amount of computational resources and time to train large datasets, making them challenging to implement. Instead of taking up such an expensive path, we can reuse the weights from a pre-trained model developed for another computer vision task with the transfer learning technique~\cite{tan2018survey}. The general idea for using transfer learning for image classification tasks is that these models’ feature maps, once trained on a large and general dataset, can be effectively used for other visual tasks. For our work, we customize the pre-trained version of MobileNetV2 trained at Google. MobileNetV2 is trained on the ImageNet~\cite{deng2009imagenet} dataset, consisting of around 1.4 million images from over 1000 classes.

MobileNetV2 architecture uses an inverted residual structure with shortcut connections between the thin bottleneck layers. This model improves image classification’s symbolic power and performance by removing the non-linearities in the narrow layers. MobileNetV2 uses two kinds of blocks: residual blocks with a stride of 1 and a block with a stride of 2 for downsizing. Both types of blocks have three layers. The first layer is 1x1 convolution with ReLU6, the second is a depthwise convolution, and the third uses 1x1 convolution without any linearity.

\begin{figure}[h]
    \centering
    \includegraphics[scale=0.16,trim={0 8.5cm 0 0},clip]{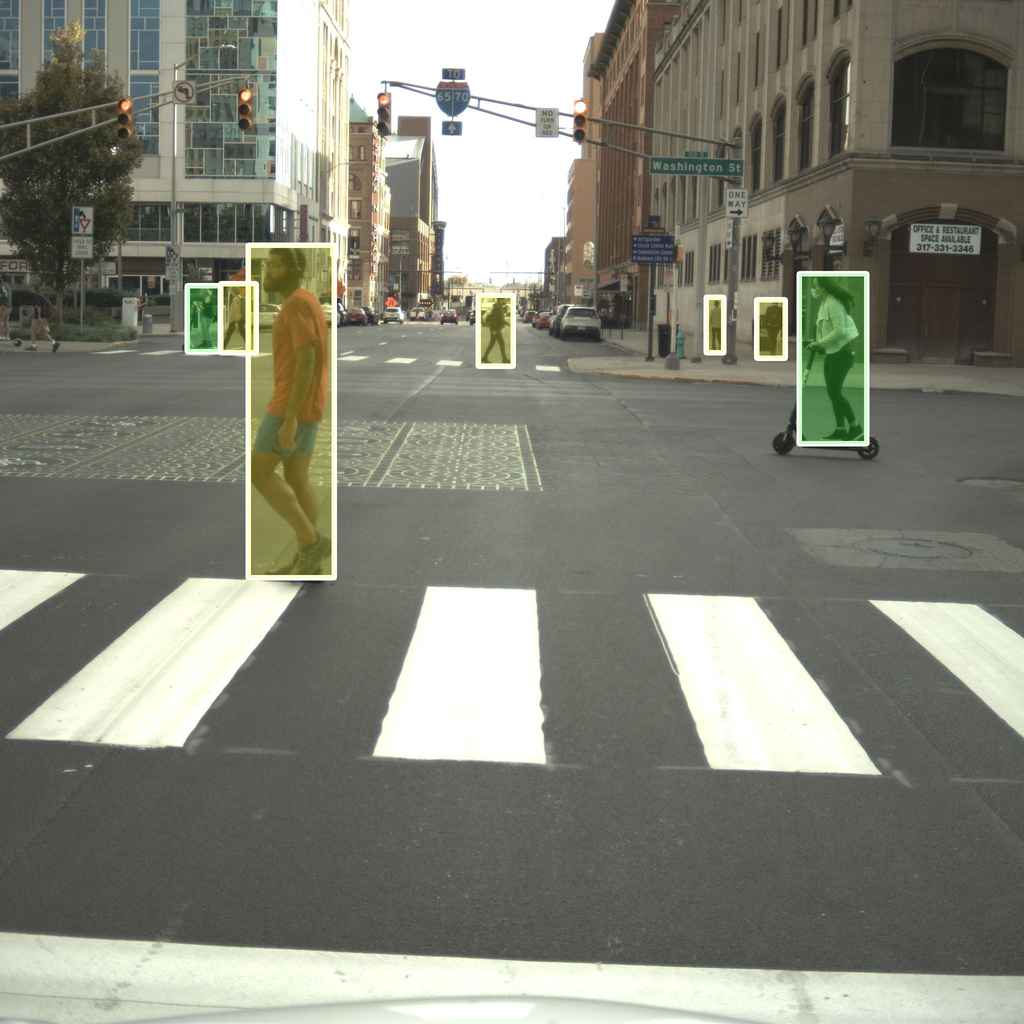}
    \caption{\textsf{MobileNetV2 outputs prediction score for each extended region; e-scooter riders are denoted by green boxes.}}
    \label{fig:step3}
\end{figure}

We choose the bottleneck layer, i.e., the last layer of MobileNetV2 before the flattening operation, to extract general features. Since the MobileNetV2 is trained on data scaled to [-1, 1], we first rescale our input data. The model acting as a feature extractor converts the 160x160x3 input image vector into a (5, 5, 1280) tensor. We perform a global pooling operation over the 5x5 spatial locations to produce a vector of length 1280 to represent every image segment. To convert these features into a raw prediction value, we use a Dense layer. The model outputs values ranging from 0 to 1, 0 for non-riders and 1 for e-scooter riders. Since the model performs a binary classification task, we use binary cross-entropy as the loss function, modeled as the equation:
\begin{equation}
    Loss = -(y\log(p) + (1-y)\log(1-p))
\end{equation}
Here, \emph{y} is the binary indicator (0 or 1) to denote the class label for observation, and \emph{p} is the predicted probability of the observation. We take these prediction values for each corresponding bounding box and generate output as shown in Fig.~\ref{fig:step3}.


\section{DATASET CREATION AND MODEL TRAINING}

\begin{figure*}[h]
    \centering
    \setkeys{Gin}{width=0.13\linewidth}
        \includegraphics{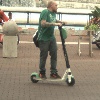}
        \includegraphics{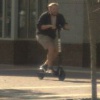}
        \includegraphics{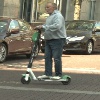}
        \includegraphics{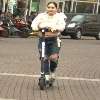}
        \includegraphics{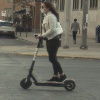}
        \includegraphics{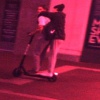}
        
        \vspace{0.13cm}
        
        \includegraphics{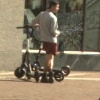}
        \includegraphics{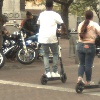}
        \includegraphics{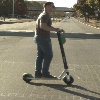}
        \includegraphics{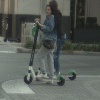}
        \includegraphics{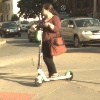}
        \includegraphics{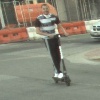}
    
    \caption{\textsf{Training examples of e-scooter riders}}
    \label{training_examples_esc}
\end{figure*}

\begin{figure*}[h]
    \centering
    \setkeys{Gin}{width=0.13\linewidth}
        \includegraphics{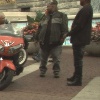}
        \includegraphics{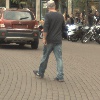}
        \includegraphics{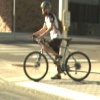}
        \includegraphics{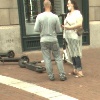}
        \includegraphics{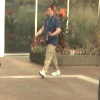}
        \includegraphics{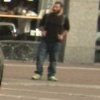}
        
        \vspace{0.13cm}
        
        \includegraphics{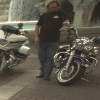}
        \includegraphics{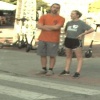}
        \includegraphics{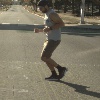}
        \includegraphics{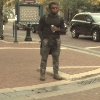}
        \includegraphics{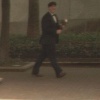}
        \includegraphics{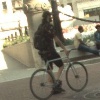}
    
    \caption{\textsf{Training examples of non riders}}
    \label{training_examples_no_esc}
\end{figure*}

\subsection{Data Collection}
We perform the e-scooter data collection using an ego vehicle with six cameras mounted on top. There are three cameras in the front (collectively covering the front, front-left, and front-right views) and three in the back (collectively covering the back, back-left, and back-right views), allowing us to capture the scene and e-scooter riders from all angles. We can capture any e-scooter riders’ entire interaction with this set-up from all angles and different distances. The data collected is representative of the real-world situation. 

We use FLIR Grasshopper-3 as our cameras, which can provide high-resolution images of 2048x2048. We operate these cameras at 10 Hz. The global shutter in Grasshopper-3 reduces the blur in pictures that can originate from fast motions. These cameras also possess high exposure ranges, allowing the system to operate in both well illuminated and low light conditions.

We separated a total of 83 unique interactions of e-scooter riders with our ego vehicle. Some of these interactions also involved multiple riders in the scene. Since we capture the entire scene from our ego vehicle, we collect the whole exchange of the e-scooter riders.

\subsection{Dataset Generation}
We separate the scenarios where our vehicle had an interaction with any e-scooter rider from the data collected. Following this, we create the e-scooter rider image dataset. Today, many existing algorithms and deep learning models, like RCNN, Fast-RCNN, YOLO, etc., can detect people in an image with high accuracy. Instead of manually labeling bounding boxes around people, we use an existing classifier (in this case, YOLOv3) to obtain the bounding boxes for persons in the frame. The YOLOv3 version trained on the MS COCO dataset classifies e-scooter riders as persons.

To label our data, we use a semi-automatic process. Firstly, we use the method described in \ref{process_step_2} to obtain the larger bounding boxes around each detection. Using these modified bounding boxes, we then extract each image segment, pixels belonging to each person, in a frame. Following this automatic process, we manually separate the images belonging to e-scooter riders from other pedestrians.

Furthermore, we use web data mining to collect some more e-scooter riders’ images. Since there is no dedicated data collection available for this use case, we could only obtain a total of 229 unique images from the web. With this, we generate the final balanced dataset for classification consisting of around 18900 total image segments, i.e., the cropped out segments from an image, from both the ‘e-scooter rider’ and ‘not an e-scooter rider’ classes. As seen in Fig.~\ref{training_examples_esc} and Fig.~\ref{training_examples_no_esc}, the training data includes e-scooter riders and non-riders from several distances and all angles.

\subsection{Model Training}
\label{model_training}
From the 83 interactions, we split our data into the train (85\%) and test (15\%) datasets. We separate a total of 18,948 image segments (9475 from e-scooter riders and 9473 from the rest) to train our e-scooter detector model. We use a different day of data collection for our test data, where we collected 12 e-scooter rider interactions. We create a balanced test dataset consisting of a total of 2497 image segments from both classes, with e-scooter riders’ exchanges coming from various angles and distances with our ego vehicle. Following this, we train MobileNetV2 to perform the binary classification task to identify e-scooter riders among pedestrians. Using a batch size of 32, we feed the data into the network and evaluate the model performance over the test dataset. 

\begin{figure*}
    \centering
    \setkeys{Gin}{width=0.24\linewidth}
        \includegraphics{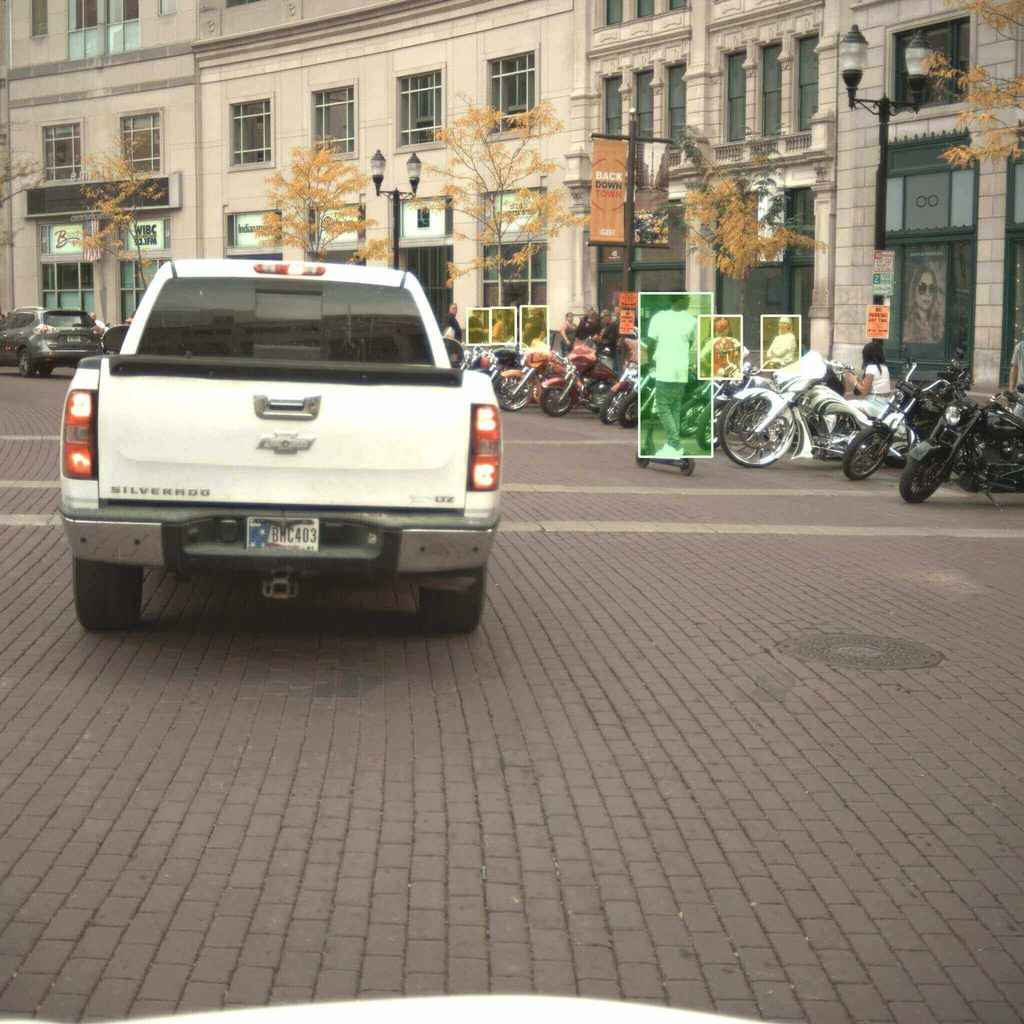}
        \includegraphics{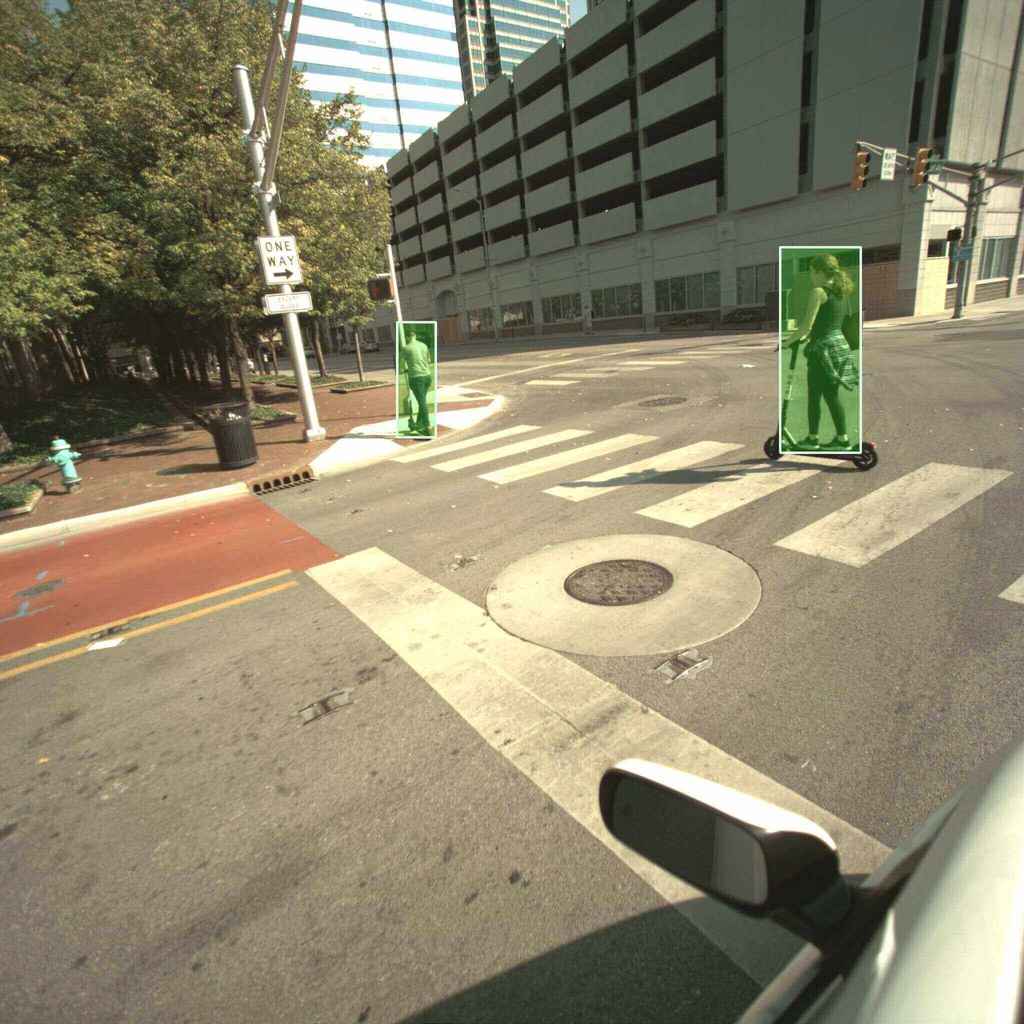}
        \includegraphics{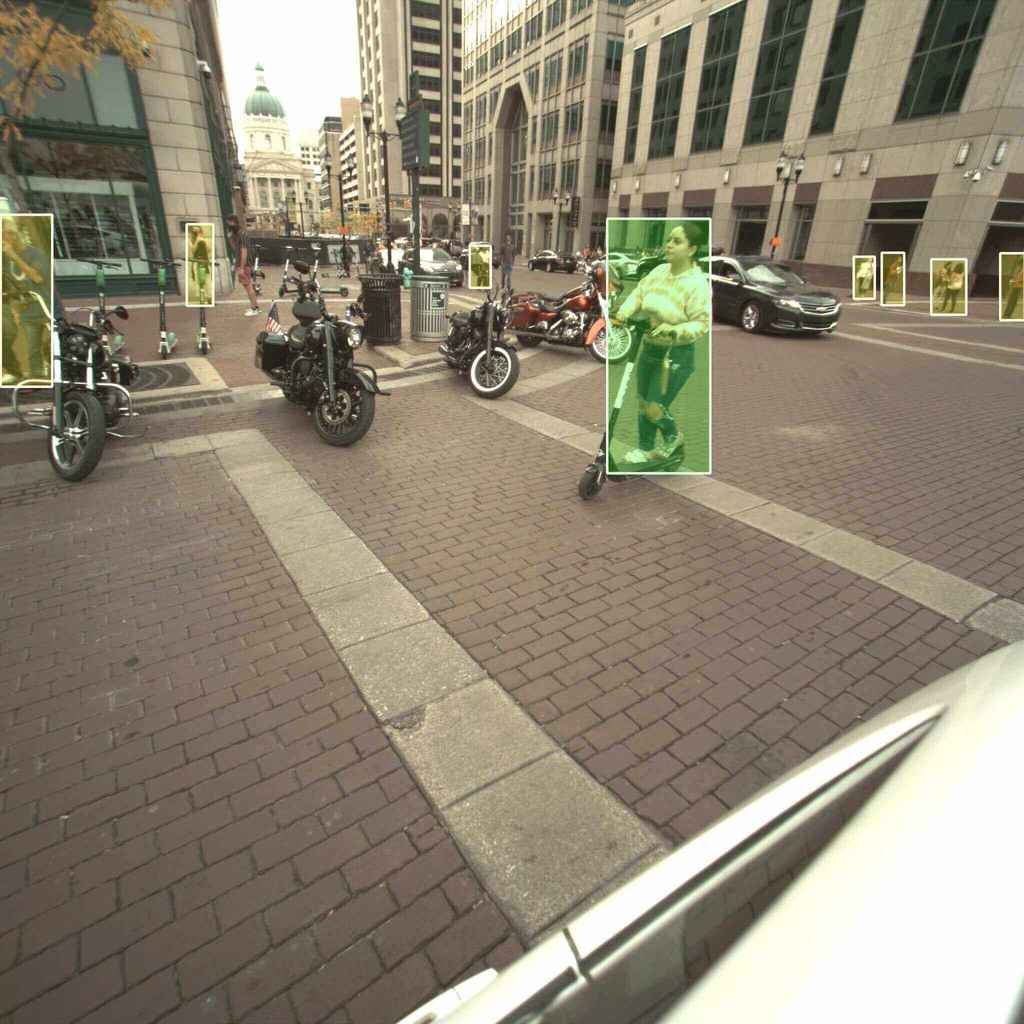}
        \includegraphics{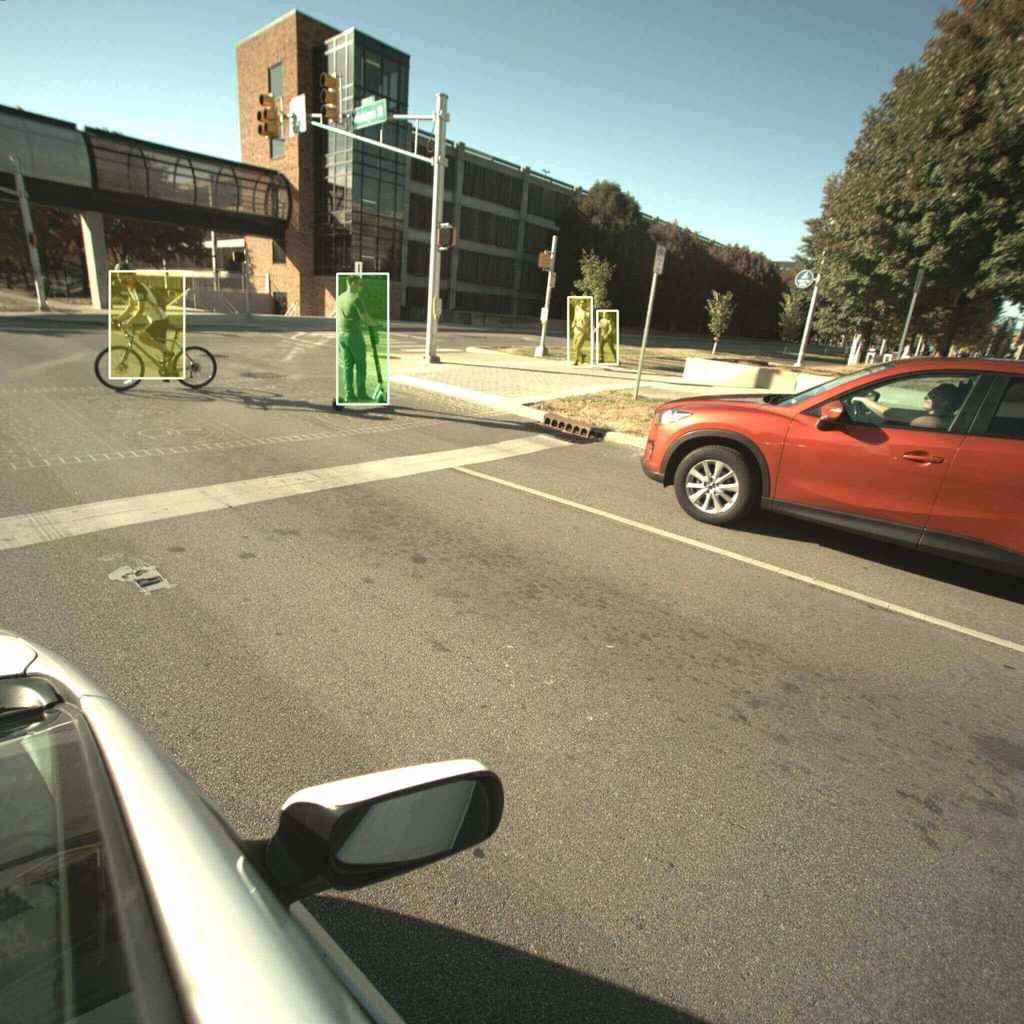}
        
        \vspace{0.13cm}
        
        \includegraphics{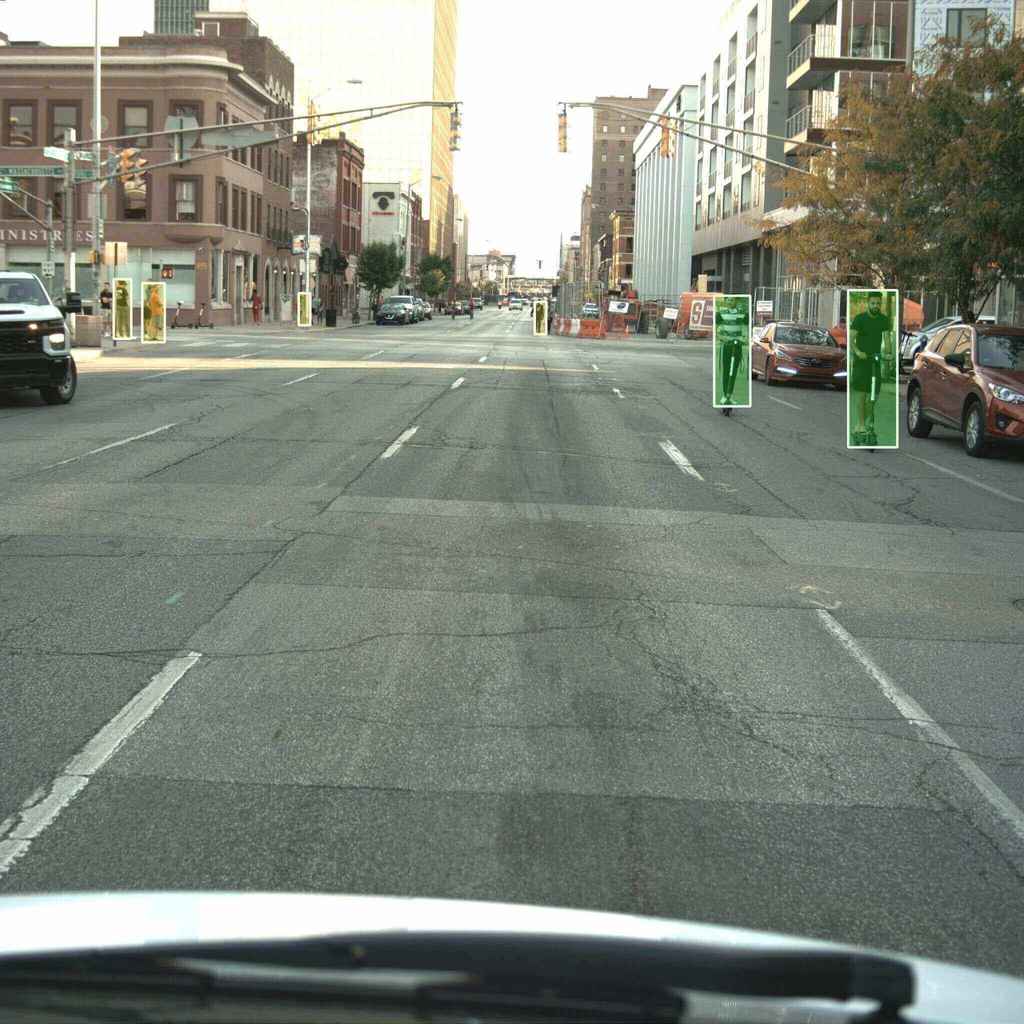}
        \includegraphics{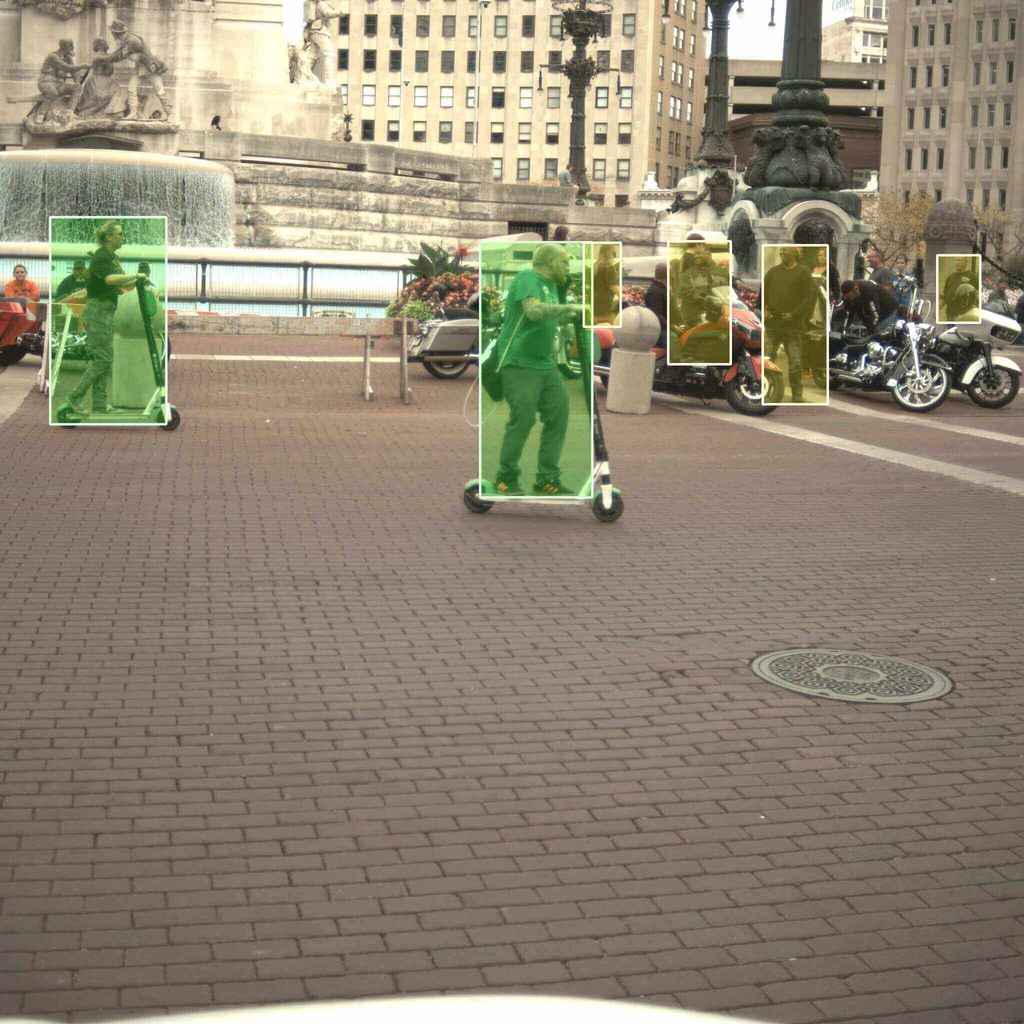}
        \includegraphics{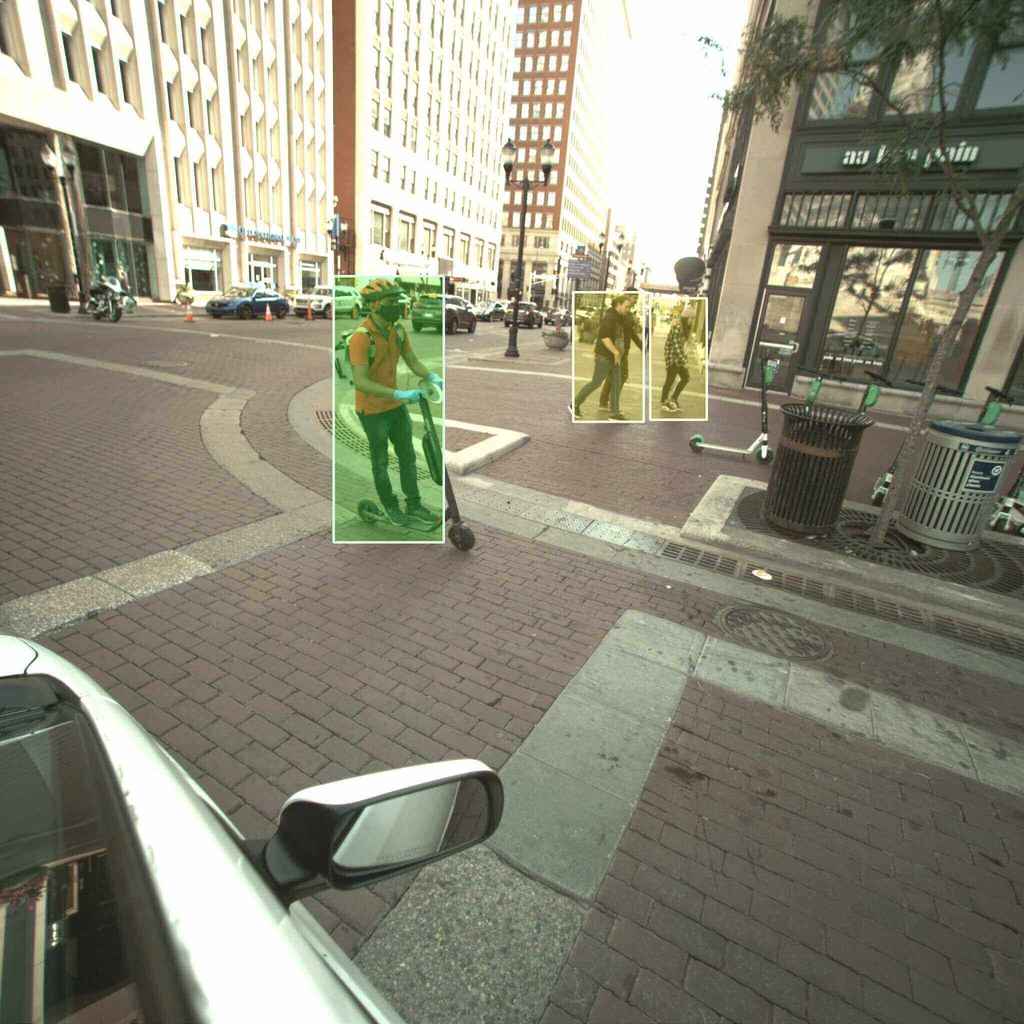}
        \includegraphics{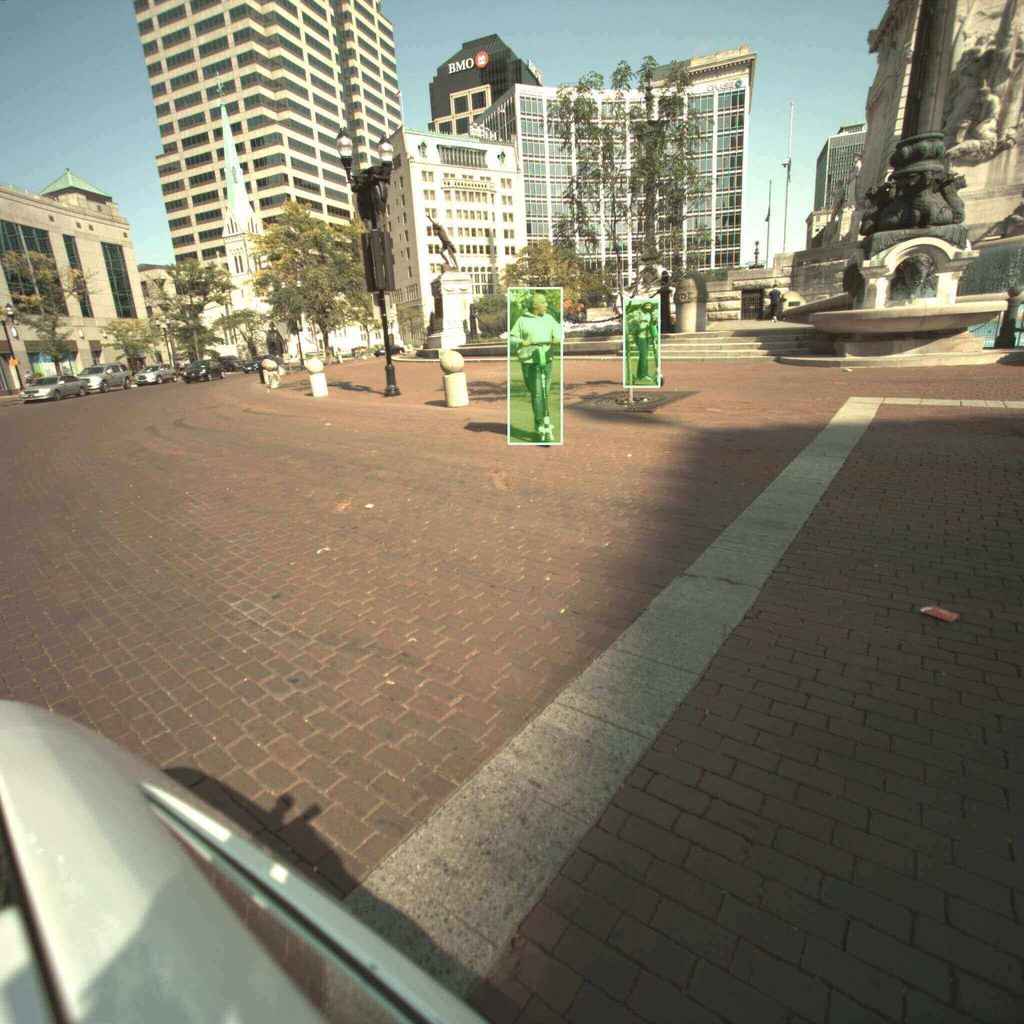}
    
    \caption{\textsf{Model performance on natural data: The model outputs bounding boxes for persons in an image with additional prediction value (1 for e-scooter rider and 0 otherwise). We show e-scooter riders in green bounding boxes and other persons in yellow bounding boxes.}}
    \label{demo_results}
\end{figure*}

After our global pooling layer in MobileNetV2, we obtain 1280 parameters. Since we add a final dense layer to generate predictions, we get 1281 trainable parameters. Initially, we keep the convolutional base of MobileNetV2 frozen before compilation and training. Freezing prevents the weights in any given layer from being updated during training. We use Adam~\cite{kingma2014adam} optimizer at the learning rate of 0.0001. We also add a dropout layer at a rate of 0.3 before our final dense layer to prevent the model from overfitting. Training this model for ten epochs gives us an accuracy of 87\% on the test data already. 

To increase the performance further, we fine-tune the model by updating the weights of some of the top layers in MobileNetV2. Since the initial layers of a deep neural network learn only simple features, we keep them frozen. Out of 154 trainable layers in MobileNetV2, we only update the weights of the top 54 layers as the features extracted in these layers would be more specific to our dataset. Thus, with this technique, the weights get tuned based on the features associated with our dataset. We train for another 15 epochs using Adam optimizer again, but with a lower learning rate of 0.00001 as the model might easily overfit now. After training, we obtain an accuracy of over 91\% and an f-measure of 0.9 on our test dataset.


\section{RESULTS}
\subsection{Metrics}
We use F1 score and ROC-AUC score to evaluate the performance of our models. F1-score is the harmonic mean of precision and recall. The precision tells us the fraction of positive predictions that were actually correct, and the recall tells us the fraction of actual positives that were identified correctly.

\[
    \emph{Precision} = \frac{\emph{True Positives}}{\emph{True Positives} + \emph{False Positives}}
\]

\[
    \emph{Recall} = \frac{\emph{True Positives}}{\emph{True Positives} + \emph{False Negatives}}
\]

\[
    \emph{F1 score} = \frac{2 \times \emph{Precision} \times \emph{Recall}}{\emph{Precision} + \emph{Recall}}
\]

Receiver Operating Characteristic (ROC) curve is a graph to show the performance of a classification model. It plots TPR (true positive rate) on the Y-axis and FPR (false positive rate) on the X-axis at different classification thresholds plotted from (0,0) to (1,1). AUC, the two dimensional area under the ROC curve, is seen as the probability of the model ranking a random positive example higher than the negative one. A value closer to 1 tells us that the model is making accurate predictions.

\subsection{Classification Performance of trained MobileNetV2 on our test dataset}
As discussed in section~\ref{model_training}, our test dataset contains 1269 image segments, i.e., cropped segments from original images of e-scooter riders and 1228 image segments of other persons. After training, MobileNetV2 could correctly predict 1137 out of 1269 e-scooter rider segments and 1131 out of 1228 segments from the other class. The trained model wrongly predicted 132 e-scooter riders as non-riders and 97 other pedestrians as e-scooter riders. So, we can see that we obtain an F-measure of over 0.91 to detect the e-scooter riders. In Table~\ref{table:training_classification_report}, we describe this performance of our trained version of MobileNetV2 on the test dataset.

\begin{table}[h]
\caption{Classification performance of MobileNetV2 on test dataset}
\centering
\begin{tabular}{c c c c c}
\hline \hline
Class Label & Precision & Recall & F-measure & Support\\ [0.5ex]
\hline
E-scooter rider & 0.92 & 0.90 & 0.91 & 1269 \\ 
Non rider & 0.90 & 0.92 & 0.91 & 1228 \\ [1ex]
\hline
 & & Accuracy & 0.91 \\ [1ex]
\hline
\end{tabular}
\label{table:training_classification_report}
\end{table}

\subsection{Overall performance of the pipeline against ground truth}
To evaluate the performance metrics of our model, we generate a test sample of 276 original raw images. Testing our entire pipeline on raw images, congruent to naturalistic scenarios, gives us a better picture of our model's performance. We randomly generate this sample from 12 different e-scooter rider interactions that we had separated as test scenarios. Our test sample consists of e-scooter riders and other pedestrians from various angles and distances in naturalistic scenarios. A total of 121 images in the test sample have at least one e-scooter rider present.

\begin{table}[!ht]
\caption{Performance of entire pipeline against the ground truth}
\centering
\begin{tabular}{c c c c}
\hline \hline
Outcome & True Positive & False Negative & Recall \\ [0.5ex]
\hline
E-scooter rider & 127 & 44 & 0.74 \\ [1ex]
\hline
\end{tabular}
\label{table:overall_performance}
\end{table}

We count the number of pedestrians and e-scooter riders present in every image from the sample through manual annotation. In our test sample, there are a total of 171 e-scooter riders and 985 pedestrians. As described in Table~\ref{table:overall_performance}, our model’s overall performance shows that it correctly identified 127 out of 171 e-scooter rider instances. Among 44 false negatives, there were 13 instances wrongly classified as pedestrians and 31 remaining ones unidentified. Our model shows a recall measure of approximately 74\%. Since our model uses YOLOv3 as the base model to detect persons, some false negative detections from YOLOv3 contribute to the overall errors (the 31 unidentified ones). This limitation is evident as the recall is not very high, and many e-scooter riders, especially far in the scene, were not detected by YOLOv3 as a person in the first place.

\subsection{Classification Performance of trained MobileNetV2 after Extended Region Extraction}

Furthermore, in Table~\ref{table:performance_over_yolo}, we evaluate the performance of only our classifier, i.e., the output of trained MobileNetV2 after candidate region selection with YOLOv3. Out of 1156 persons in the sample, YOLOv3 can detect 762 of them. There are a total of 140 e-scooter riders out of these 762 people, and our model predicts 127 of them correctly. The remaining 622 detections from YOLOv3 are other pedestrians, and our model can correctly predict 616 of them.

\begin{table}[h]
\caption{Performance analysis of MobileNetV2 on top of YOLOv3}
\centering
\begin{tabular}{c c c c}
\hline \hline
Class Label & Precision & Recall & F-measure \\ [0.5ex]
\hline
E-scooter rider & 0.95 & 0.91 & 0.93 \\ 
Non rider & 0.98 & 0.99 & 0.98 \\ [1ex]
\hline
 & & Accuracy & 0.975 \\ [1ex]
\hline
\end{tabular}
\label{table:performance_over_yolo}
\end{table}

These results show that we have precision and recall both over 0.9 for both classes. With an F-measure of around 0.9, our trained version of MobileNetV2 can differentiate between the two classes well. The overall accuracy of MobileNetV2 on this test sample is around 97.5\%. We show some of our outputs in Fig.~\ref{demo_results}.


\section{CONCLUSIONS}

This paper proposes a classification approach and introduces a novel benchmark dataset for e-scooter riders in naturalistic traffic scenes. We exploit an existing classifier, YOLOv3, and train MobileNetV2 to distinguish between e-scooter riders and other pedestrians. We can successfully differentiate between an e-scooter rider and any other pedestrians on the road with this vision approach. Our system produces highly accurate results with precision and recall over 0.9. With this detection system, we aim to help the autonomous driving community study and understand e-scooter riders’ behavior. 

To improve our current model, we can train using more data from the cases involving overlapping of e-scooter riders and pedestrians. When the e-scooter rider is far and facing away from the camera, the body of the e-scooter hides, making it difficult for the model to classify correctly across frames. We also wish to study particular cases like these. Moreover, in our future work, we aim to build a single neural network as a multiclass classification system to identify various vulnerable road users who use other vehicles. 

\addtolength{\textheight}{-12cm}   








\bibliographystyle{unsrt}

\end{document}